%% file: emnlp2020.tex
\documentclass[11pt,a4paper]{article}
\usepackage{times,latexsym}
\usepackage{url}
\usepackage[T1]{fontenc}
\usepackage[]{emnlp2020}

\usepackage{hhline}
\usepackage{booktabs}
\usepackage{array}
\usepackage[normalem]{ulem}
\usepackage{hyperref}
\usepackage{tabularx} 
\usepackage{mathbbol} 
\usepackage{xfrac,nicefrac} 
\usepackage[super]{nth} 
\usepackage[export]{adjustbox} 
\usepackage{float}

\DeclareUrlCommand{\bulurl}{} 

\usepackage{xcolor} 

\def\@fnsymbol#1{\ensuremath{\ifcase#1\or *\or \dagger\or \ddagger\or
   \mathsection\or \mathparagraph\or \|\or **\or \dagger\dagger
   \or \ddagger\ddagger \else\@ctrerr\fi}} 
\newcommand{\ssymbol}[1]{^{\@fnsymbol{#1}}} 

\usepackage{tabularx,ragged2e}
\newcolumntype{L}[1]{>{\RaggedRight\arraybackslash%
     \hspace{0pt}
     \hsize=#1\hsize}X}
\usepackage{amsmath,bm}
\usepackage{amssymb}
\usepackage{mathtools}
\usepackage{arydshln}
\usepackage[flushleft]{threeparttable}
\usepackage{soul}
\usepackage{url}
\usepackage{paralist}
\usepackage{rotating,makecell}
\usepackage{tabularx}
\usepackage{array}
\usepackage{lipsum}
\usepackage{caption}

\usepackage{tabu}
\usepackage{comment}

\usepackage{ragged2e} 
\usepackage{xparse}
\NewExpandableDocumentCommand\mcx{O{1}m}
    {\multicolumn{#1}{>{\Centering\small\bfseries\hsize=#1\hsize}X}{#2}}
\usepackage{ragged2e}
\usepackage{siunitx}

\usepackage{xr}

\usepackage{color}
\usepackage{enumitem}
\usepackage[linecolor=orange]{todonotes}
\newcommand{\hlc}[2][yellow]{{\sethlcolor{#1}\hl{#2}}}
\definecolor{orange}{RGB}{255, 50, 190}
\definecolor{green}{RGB}{228, 249, 189}
\definecolor{Green}{RGB}{141, 213, 134}
\definecolor{pink}{RGB}{253, 221, 210}
\definecolor{pink}{RGB}{255, 204, 204}

\definecolor{light-blue}{RGB}{193,225,236}
\definecolor{yellow}{RGB}{255, 248, 194}
\definecolor{darkpastelgreen}{rgb}{0.01, 0.75, 0.24}
\definecolor{cerise}{rgb}{0.87, 0.19, 0.39}
\definecolor{celestialblue}{rgb}{0.29, 0.59, 0.82}
\definecolor{cerulean}{rgb}{0.0, 0.48, 0.65}
\definecolor{darkgray}{rgb}{0.66, 0.66, 0.66}
\definecolor{gray}{rgb}{0.75, 0.75, 0.75}

\usepackage{multirow}
\usepackage{graphicx}

\usepackage{xspace,mfirstuc,tabulary}

\fi

\usepackage{microtype}

\aclfinalcopy 
\def\aclpaperid{3028} 

\title{Modeling Protagonist Emotions for Emotion-Aware Storytelling}

\author{Faeze Brahman \\
  University of California\\ Santa Cruz  \\
  \texttt{fbrahman@ucsc.edu} \\\And
  Snigdha Chaturvedi \\
  University of North Carolina \\ Chapel Hill \\
  \texttt{snigdha@cs.unc.edu} \\
  }

\date{}

\begin{document}
\maketitle
\begin{abstract}

Emotions and their evolution play a central role in creating a captivating story. In this paper, we present the first study on modeling the emotional trajectory of the protagonist in neural storytelling. We design methods that generate stories that adhere to given story titles and desired \textit{emotion arcs} for the protagonist. 
Our models include Emotion Supervision (EmoSup) and two Emotion-Reinforced (EmoRL) models. The EmoRL models use special rewards designed to regularize the story generation process through reinforcement learning.
Our automatic and manual evaluations demonstrate that these models are significantly better at generating stories that follow the desired emotion arcs compared to baseline methods, without sacrificing story quality.

\end{abstract}

\section{Introduction}
\label{intro}

Stories are an integral part of human culture. 
They allow us to express emotions, share knowledge, and to shape our perspective of the world~\cite{Mckee2003}.
Stories are made interesting through emotions that connect the characters, their motivations, goals, and achievements~\cite{vonnegut}.

Cognitive scientists have pinpointed the central role of emotions in storytelling~\cite{parkinson,hogan2011literature}.
Early automatic storytelling systems based on symbolic planning also showed that addressing character emotions for plot construction resulted in more diverse and interesting stories 
\cite{theune2004,perez2007,mendez2016}. However, these studies were rule-based and limited to small-scale data. The advent of deep learning has shifted computational storytelling efforts towards neural methods~\cite{Martin:18,Yao:19}.
However, despite the broad recognition of its importance, neural story generation methods have not explored the modeling of emotional trajectory.

\input{figures/example.tex}

In this paper, we present the first study to take into account the emotional trajectory of the protagonist in neural story generation. 
Research in cognitive science has shown that while comprehending narratives, readers closely monitor the protagonist’s emotional states~\cite{Komeda2006TheEO, Gernsbacher:92}. However, emotions experienced by the protagonist might differ 
from the general emotions expressed in the story.  For example, the general emotion of \textit{``My boss was very angry and decided to fire me.''} is \textit{anger}, but the narrator's emotional reaction would be to feel \textit{upset}. 
At any point in a story, we represent the protagonist's emotions using a set of \textit{basic emotions}. 
The theory of basic emotions is well-accepted in psychology, but there is little consensus about the precise number of basic emotions. \citet{plutchik} proposed $8$ primary emotions, and \citet{ekman1992argument} first proposed $7$ and then changed to $6$ basic emotions. Following recent theories~\cite{jack2014dynamic, gu2016neuromodulator}, we choose \textit{anger, fear, joy}, and \textit{sadness}, to describe the protagonist's emotions. We additionally include \textit{neutral} to account for cases with no strong emotions. We refer to these $5$ emotions as \textit{basic emotions}.

Moreover, emotions 
evolve within a narrative. 
For modeling the evolving emotions of the protagonist, we define an \textit{emotion arc} for a story.
Our definition is inspired by Prince's change-of-state formalization~\cite{Prince}, which asserts that stories are about change. According to this theory, a story has three components: a starting state; an ending state; and events that translate the starting into the ending state. Motivated by this, 
we define the \textit{emotion arc} as a sequence of three \textit{basic emotions} that describe the starting, body, and ending emotional states of the protagonist. 

Given a story title and the emotion arc of the protagonist as inputs, our goal is to generate a story about the title that adheres to the given emotion arc. Fig.~\ref{example} shows an example story generated by our model, where the protagonist's emotion evolves from \hlc[green]{\textit{joy}} to \hlc[pink]{\textit{anger}} and then \hlc[light-blue]{\textit{sadness}}. \par 

To address this problem, we present three models based on GPT-2~\cite{radford2019language} that incorporate the protagonist's emotion arc as a controllable attribute while preserving content quality: an Emotion Supervision (EmoSup), and two Emotion-Reinforced (EmoRL) models based on reinforcement learning. The EmoRL models use two Emotion-Consistency rewards, \textsc{Ec-Em} and \textsc{Ec-Clf}. 
\textsc{Ec-Em} uses semantically enhanced emotion matching to encourage the model to adhere to the given emotion arc. It infers the protagonist's emotions in the generated stories using \textit{Commonsense Transformers},  $\mathbb{COMET}$~\cite{Bosselut2019COMETCT}. \textsc{Ec-Clf} achieves the same goal using a classifier to infer the protagonist's emotions.

In the absence of a training corpus of stories labeled with the protagonist's emotions, we automatically annotate a large-scale story corpus using $\mathbb{COMET}$. Our automatic and manual evaluations show that 
our 
models can not only express the desired emotion arcs but also produce fluent and coherent stories. Our contributions are: 
\begin{itemize}[noitemsep,topsep=0.3pt]
    \item We present the first study on modeling the \textit{emotion arc} of the protagonist for neural story generation.
    \item We propose 
    two Emotion-Consistency rewards designed to enforce the desired emotion arcs 
    using reinforcement learning. 
    \item We track the protagonist's emotions in a story (1) using a commonsense knowledge model based pipeline; and (2) through an emotion classifier trained using transfer learning from out-of-domain data.
    \item We empirically demonstrate that our models can effectively generate stories that follow the desired emotion arc. We also illustrate how these models can find novel applications. 
\end{itemize}

\section{Related Works}

Early story generation systems relied on symbolic planning~\cite{Lebowitz:87,Perez:01,Porteous:09,Reidl:10} or case-based reasoning~\cite{Gervas:05}. Although these systems 
could ensure long-term 
coherence, they could only operate in predefined domains and required manual engineering. These problems have been somewhat alleviated by recent seq2seq storytelling models~\cite{roemmele2016writing,jian:17}, 
some of which are based on intermediate 
representations~\cite{Martin:18, xu:18, fan-etal-2018-hierarchical, Yao:19, fan:19}.

Recent approaches have also used large-scale language models (LMs) based on Transformers~\cite{Vaswani:17}, such as  
GPT-2~\cite{radford2019language}. 
Being trained on large amounts of data, these models can generate highly fluent text and find applications in story generation~\cite{qin-etal-2019-counterfactual,Guan2020Story} and dialogue systems~\cite{budzianowski-vulic-2019-hello, wolf}. However, they lack the ability to dictate any auxiliary objective for the generated text, such as expressing specific attributes.\par

To address this, approaches such as conditional training or weighted decoding have been proposed to control different properties of the generated text such as sentiment, tense, speaker style, and length~\cite{kikuchi-etal-2016-controlling, pmlr-v70-hu17e, ghazvininejad-etal-2017-hafez, wang-etal-2017-steering, fan-etal-2018-controllable}. 
\citet{ijcai2019-829} use reinforcement learning for generating goal-driven story plots, which are sequences of event tuples. 
\citet{PlugPlay} propose 
PPLM, which uses 
an attribute classifier to steer text generation without further training the LM. \par 

More closely related to our work is generating automatic responses with a specific sentiment or emotion~\cite{zhou17, huang-etal-2018-automatic, zhou-wang-2018-mojitalk, song-etal-2019-generating}. 
Modeling characters \cite{bamman:2013,bamman:2014, vala:2015, Kim2019, krishnan:2015}, their relationships~\cite{ agarwal-etal-2013-sinnet, iyyer:2016, ChaturvediSDD16, SrivastavaCM16, ChaturvediID17}, and sentiment trajectory (\citet{chaturvedi-etal-2017-story}, \citet{zhou:19}) have been shown to be useful for story understanding, in general. However, there are limited works on incorporating characters or sentiment for story generation. Previous work model characters but not sentiment~\cite{clark-etal-2018-neural,char-centric}. \citet{scaffold:20} is a contemporary work that incorporates sentiment while ``filling in'' a narrative. \citet{peng-etal-2018-towards} and \citet{luo-etal-2019-learning} 
control the overall sentiment for story ending generation. These works are limited to coarse-grained
sentiments and/or only target the ending sentence. Instead, we model the emotional trajectory of the protagonist as the story progresses, which is more central to storytelling than the overall sentiment.

\section{Emotion-aware Storytelling}
We first 
explain how our models can track the protagonist's emotional trajectory (\S\ref{comet}). We then define the problem statement (\S\ref{prob}) and followed by an introduction to our base storytelling model (\S\ref{base-model}), which is used as the backbone of our three proposed models (\S\ref{lft} and \S\ref{rl-models})\footnote{Code at: \url{https://github.com/fabrahman/Emo-Aware-Storytelling}}.

\subsection{Tracking Protagonist's Emotions \label{comet}}

In this work, we define the \textit{protagonist} as the 
most frequently occurring character in a narrative~\cite{prominent:85}. Our two rewards for the EmoRL models (\S\ref{rl-models}) need to track the protagonist's emotions to guide the generation. For this, we obtain their emotions at various stages in the story using one of the following two approaches:

\noindent\textbf{Commonsense Transformer}\space\space\space Our \textsc{Ec-Em} reward uses a commonsense knowledge model to reason about the implicit emotional states of the protagonist. We use $\mathbb{COMET}$~\cite{Bosselut2019COMETCT}, a knowledge base construction model trained on ATOMIC \textit{if-then} knowledge triples~\cite{atomic}. It contains information about everyday events and their causes and effects. Given an event and a relation, $\mathbb{COMET}$ can generate commonsense inferences about the relation. 
For tracking emotions, we use relations \texttt{xReact} and \texttt{oReact} that correspond to emotional reactions to events  
(more details on this in \S\ref{data-prep}). \par

\noindent\textbf{Emotion Classifier}\space\space\space
Our \textsc{Ec-Clf} reward captures the protagonist's emotions using an emotion classifier. For this, we adapt the pre-trained BERT$_{\textit{large}}$ for multi-label classification over $5$ \textit{basic emotions}: \textit{anger, fear, joy, sadness,} and {neutral}. Following \citet{devlin-etal-2019-bert}, we use a fully-connected layer over the final hidden representation corresponding to the special classification token (\texttt{[CLS]}). 
We train this classifier in two steps. 

First, we train this classifier on a human-annotated dataset for emotion identification in tweets~\cite{mohammad-etal-2018-semeval}, consisting of $6,857$ tweets, with binary labels for $11$ emotions, among which we only focus on our \textit{basic emotions}. On this dataset, the classifier achieves better or comparable performance to state-of-the-art results~\cite{kant2018practical} (see Appendix \ref{appendix:clf-res} for detailed results). \par

Next, in order to identify the protagonist's emotions from a given story-text, we further fine-tune the classifier on story training data that is automatically annotated with the protagonist's emotions using the pipeline described in \S\ref{data-prep}. To evaluate the classifier, 
we obtain manual annotations for the protagonist's emotions on Amazon Mechanical Turk for a subset of $50$ randomly selected stories ($250$ sentences) from our story corpus. Each sentence was annotated by $3$ judges. Workers agreed with our emotion classifier $70\%$ of the time (random agreement would be $20\%$). See Appendix \ref{appendix:manual-annot} for more details about these annotations.

\subsection{Problem Statement \label{prob}}

We formulate the emotion-aware storytelling task as follows: given a story title as a sequence of tokens $\bm{t}{=}\{t_1, t_2, ...,t_m\}$, and an emotion arc for the protagonist as a sequence of \textit{basic emotions} $\bm{a}{=}\{e_1, e_2, e_3\}$, the task is to generate a story as a sequence of tokens $\bm{y}{=}\{y_1, y_2, ...,y_n\}$ that adheres to the title and emotion arc.\par

\subsection{Transformer-based Storytelling Model \label{base-model}}
Our models are built upon a base storytelling model that can generate a story consistent with a given prompt (e.g., title).
We choose GPT-2 (medium)~\cite{radford2019language} because our initial experiments demonstrated that it outperforms other state-of-the-art story generation models, in general (\S \ref{base-res}). GPT-2 uses multiple Transformer blocks of multi-head self-attention and fully connected layers (the left box in Fig.~\ref{model-arch}).  Since it was trained on a broad range of domains, we fine-tune it on a dataset of stories (\S \ref{data-prep}) by minimizing the negative conditional log-likelihood: 
\vspace{-0.2cm}
\begin{equation}
\begin{aligned}
    \mathcal{L}_{ML} = - \sum _{i=m}^{m+n} \log p(y_{i}|y_{<i},\bm{t})\\ 
    p(y_{i}|y_{<i},\bm{t}) = \mathrm{softmax}(h_{i}^{L}W^T) \\ 
    h_{i}^l = \mathrm{block}(h_{<i}^{l-1}), l \in [1,L] \\
    h_i^0 = W_i + P_i
\end{aligned}
\end{equation}


\noindent where $m$ and $n$ denote the number of tokens in the title and story respectively. $h_i^l$ is the $l$-th layer's output at the $i$-th position computed through transformer block with the masked multi-head self attention mechanism, and $h_i^0$ is a summation of \textit{token embedding} $W_i$ and \textit{position embedding} $P_i$ for the $i$-th token. $y_{<i}$ indicates left context. 

\subsection{Emotion Supervision (EmoSup) Model  \label{lft}}
The underlying idea behind our Emotion Supervision (EmoSup) model is to provide the emotion arc as an additional input similar to conditional training~\cite{fan-etal-2018-controllable, kikuchi-etal-2016-controlling}. Specifically, 
each title has the corresponding emotion arc prepended at the beginning, separated by a delimiter token (<\$>). This way, emotion arcs receive special treatment~\cite{kobus-etal-2017-domain}, since they are propagated to all of the story and the model learns to maximize $p(y_{i}|y_{<i},\bm{t}, \bm{a})$.\par

\subsection{Emotion-Reinforced (EmoRL) Models \label{rl-models}}
The emotion arc guides the generation in EmoSup as an initial input.  
However, we want to continually supervise the model \textit{during} the generation process. This motivates us to use a reinforcement learning framework. To deal with exposure bias, many previous works have optimized the evaluation measures (e.g., BLEU, ROUGE, CIDEr) as rewards~\cite{Rennie2017SelfCriticalST, PaulusXS18}. Here, we propose two Emotion Consistency rewards, \textsc{Ec-Em} and \textsc{Ec-Clf}, which optimize adherence to the desired emotion arc.\par

\noindent\textbf{\textsc{Ec-Em} Reward}\space\space\space This reward quantifies the alignment of the emotion arc of the generated story to the desired arc using the commonsense knowledge model, $\mathbb{COMET}$. For an $N$-sentence-long generated story, we use $\mathbb{COMET}$ to obtain the protagonist's emotional reaction for each sentence, resulting in a sequence of emotion-phrases $\bm{a}^g{=}\{g_1,g_2,...,g_N\}$\footnote{$N{=}5$ for our dataset. Also, $\mathbb{COMET}$'s outputs, $g_i$s, are phrases representing emotional reactions. Details on obtaining emotional reactions during training are provided in Appendix \ref{appendix:label-training}.}.
We then define the reward as a modified Levenshtein distance~\cite{levenshtein1966bcc} between the generated reactions $\bm{a}^g$ and the desired emotion arc $\bm{a}^*{=}\{e_1,e_2,e_3\}$. This modification   allows only two operations: (1) Deletion of an emotion-phrase (in $\bm{a}^g$), and (2) Replacement of an emotion-phrase  with a basic emotion at a cost proportional to semantic similarity between the two (e.g., \textit{happy to help} and \textit{joy}). 
Semantic similarities are computed using cosine similarity between the averaged GloVe embeddings~\cite{pennington2014glove}~\footnote{We experimented with contextualized representations from pre-trained BERT but opted for static embeddings because (1) they yielded better results, and (2) enabled us to focus on semantics of the individual emotion tokens/phrases rather than context-sensitive representations.}. The reward is defined as:
\vspace{-0.2cm}
\begin{equation}
    r_{em} = \mathrm{lev}(\bm{a}^g, \bm{a}^*) 
\end{equation}
\noindent where $\mathrm{lev}$ denotes the modified Levenshtein distance. We refer to the model that uses this reward as \textbf{\textsc{Rl-Em}}. 

\begin{figure}
  \includegraphics[scale=0.37]{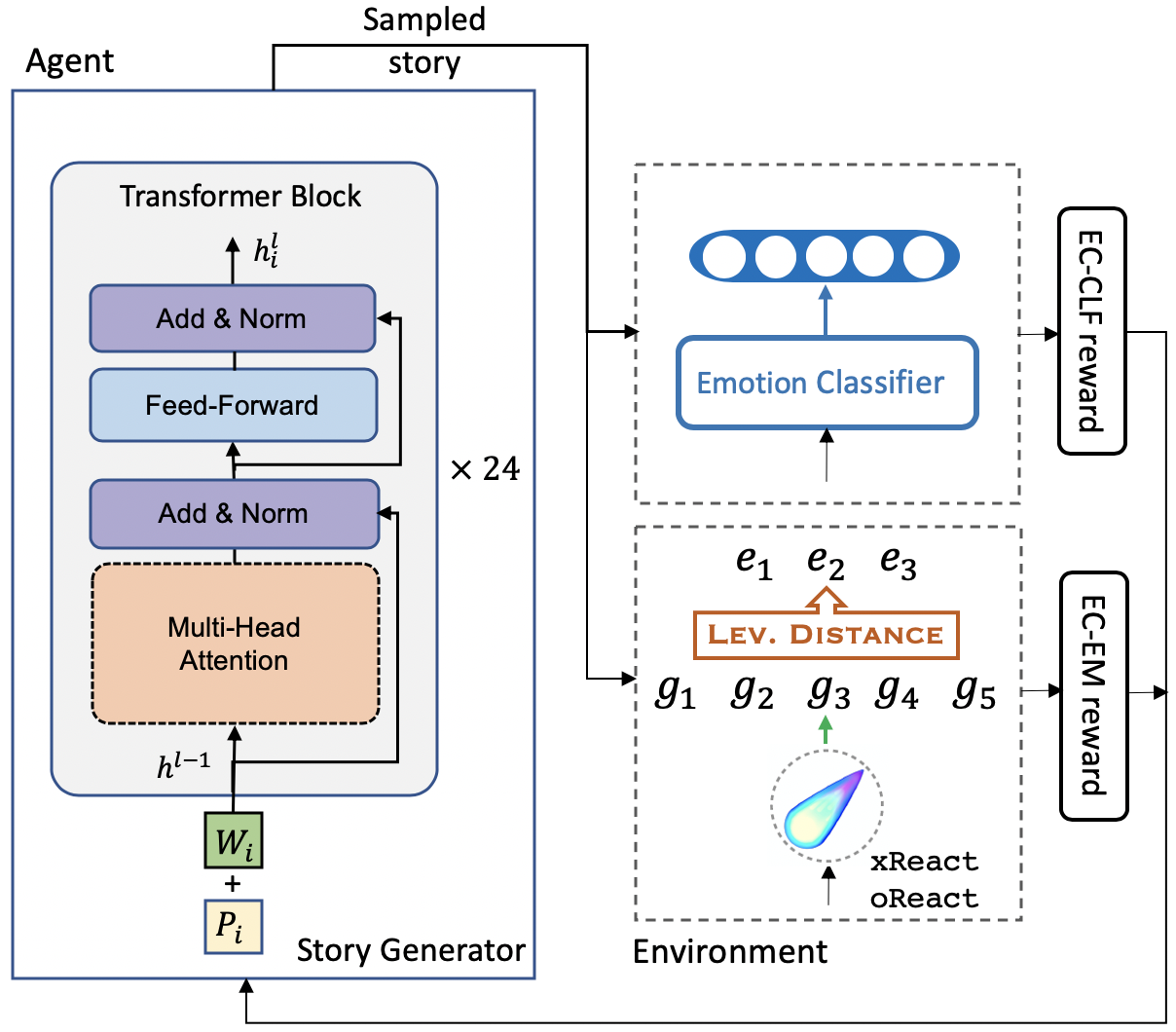} 
  \caption{Transformer block architecture (left) and emotion-reinforced storytelling framework (right)}
  \label{model-arch}
\end{figure}

\noindent\textbf{\textsc{Ec-Clf} Reward}\space\space\space This reward infers the protagonist's emotions in a given text using our 
emotion classifier (\S\ref{comet}). 
We first divide the generated story into segments: beginning, body, and ending\footnote{We generate 5-sentence long stories similar to our training corpus and segment them into the beginning, body, and ending in 1:3:1 ratio.}. Then, for each segment, we use the classifier to obtain the probability 
of the desired emotion. 
The reward is defined as the probabilities of the desired emotions averaged across the segments:
\vspace{-0.2cm}
\begin{equation}
    r_{clf} = \frac{1}{k}\sum_{j=1}^{k} p_{\mathrm{clf}}(e^*_j|\bm{x}_j)   
\end{equation}

\noindent where $k$ is the number of tokens in the emotion arc (here, $k{=}3$), and $e^*_j$ denotes the desired emotion for $j$-th segment $\bm{x}_j$. We refer to the model that uses this reward as \textbf{\textsc{Rl-Clf}}.

 \begin{figure}[t] 
  \includegraphics[scale=0.54]{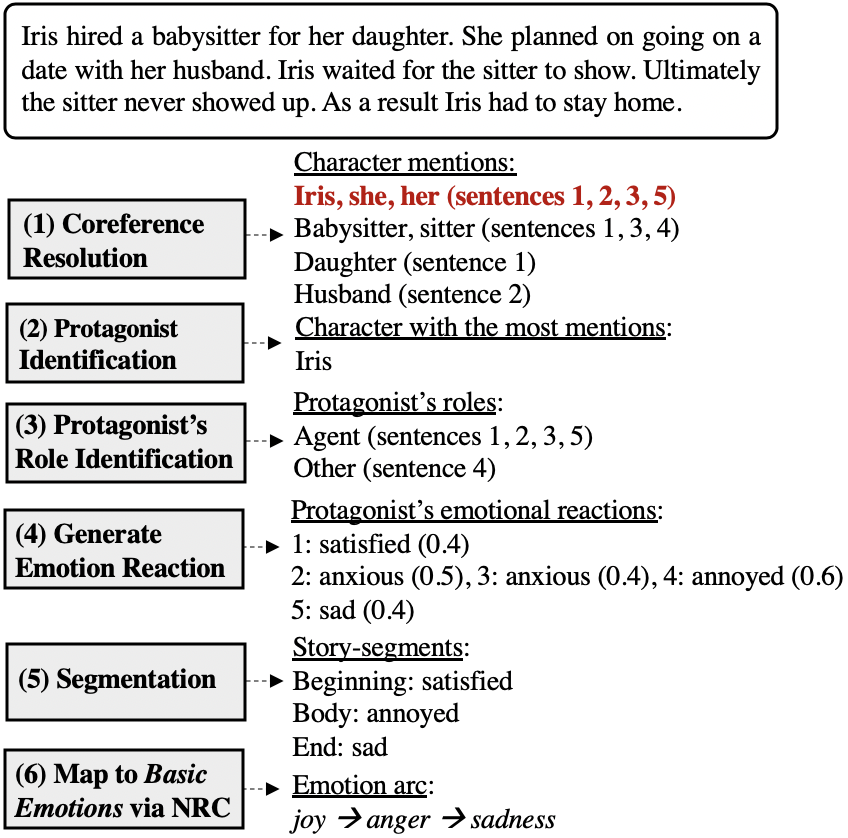} 
  \caption{Annotation pipeline for \textit{emotion arc}.}
  \label{annotation-pipe}
\end{figure}

\noindent\textbf{Policy Gradient}\space\space\space For training, we use the REINFORCE algorithm~\cite{reinforce} to learn a generation policy $p_{\theta}$ of the storytelling model with parameters $\theta$. Here, the model generates a sample story $\bm{y}^s$ from the model's output distribution, and the goal is to minimize the negative expected reward, which is approximated by:
\vspace{-0.3cm}
\begin{equation}
\small{
   \mathcal{L}_{RL} = - (r({\bm{y}}^s) - r(\hat{\bm{y}}))\sum _{i=k+m}^{k+m+n} \log p_{\theta}(y^s_{i}|y^s_{<i})
   }
\end{equation}

We follow the self-critical training approach~\cite{Rennie2017SelfCriticalST}, and take the reward of the greedily decoded story $\hat{\bm{y}}$ as the baseline reward $(r(\bm{\hat{y}}))$. This ensures that with better exploration, the model learns to generate stories $\bm{y^s}$ with higher rewards than the baseline $\hat{\bm{y}}$. Optimizing only with the RL loss mentioned above using the emotion-consistency rewards may increase the expected rewards, but at the cost of 
 fluency and readability of the generated story. Therefore, we optimize the following mixed loss~\cite{PaulusXS18}:
\vspace{-0.2cm}
\begin{equation}
    \mathcal{L}_{mixed} = \gamma \mathcal{L}_{RL} + (1- \gamma)\mathcal{L}_{ML}
\end{equation}
where $\gamma$ is a hyper-parameter balancing the two loss functions. Our emotion-reinforced storytelling framework is depicted in Fig.~\ref{model-arch}.

\section{Experimental Setup}

\subsection{Dataset and Annotation Pipeline \label{data-prep}}

We use the \textit{ROCStories} corpus~\cite{Mos:16} for our experiments. It contains $98,162$ five-sentence stories, 
designed to have a clear beginning and ending
, thus making it a good choice for our emotion-aware storytelling task. We held out $10\%$ of the stories each for validation and test sets, respectively. 

For training our models, we need stories annotated with emotion arcs of the protagonists. We annotated the stories in our dataset automatically using the multi-step annotation pipeline shown in Fig.~\ref{annotation-pipe}. In step 1, we identify all characters and their mentions in a story using coreference resolution. In step 2,  we identify the character with the most mentions as the \textit{protagonist} (e.g., `Iris' who is mentioned in $4$ sentences). Then, in step 3, in each sentence of the story, we identify the protagonist's role as \textit{Agent} or \textit{Other} using its dependency parse\footnote{We use AllenNLP~\cite{gardner-etal-2018-allennlp} for coreference resolution and dependency parsing: \url{https://github.com/allenai/allennlp}}. The protagonist is an \textit{Agent} if he/she is the subject of the main verb in the sentence and \textit{Other} otherwise (e.g., Iris's role is \textit{Other} in sentence 4 and \textit{Agent} in all other sentences).  Next, in step 4, we obtain the emotional reaction of the protagonist in each sentence using $\mathbb{COMET}$. Given a context, $c$, and relation type, $r$, $\mathbb{COMET}$ can yield the emotional reaction of the \textit{Agent} (r=\texttt{xReact}) and \textit{Others} (r=\texttt{oReact}). Depending on the protagonist's role in the sentence, we use the appropriate relation to get their emotional reaction, $g$, and $\mathbb{COMET}$'s  confidence in the prediction, ${\varphi}_g$.
In sentences without an explicit mention of the protagonist, his/her role is assigned as \textit{Other}, and we use \texttt{oReact} since the event in that sentence will affect all characters of the story, including the protagonist (e.g., sentence 4 in Fig.~\ref{annotation-pipe}).

Step 4 gives the protagonist's emotions for each sentence of the story, but the emotion arc has to represent them for the three segments: beginning, body, and end. The stories in our corpus are $5$-sentence long, and following previous work on this corpus~\cite{chaturvedi-etal-2017-story}, we segment them in 1:3:1 ratio. 
For the protagonist's emotion in the body (middle $3$ sentences), we take the 
emotion of the sentence in which $\mathbb{COMET}$ was most confident (e.g., `annoyed' for the body of the running example in Step 5).

Note that since $\mathbb{COMET}$'s outputs, $g$s, are open-ended emotion-phrases, in step 6, we need to map these phrases to one of the $5$ \textit{basic emotions} using NRC Affect Intensity Lexicon~\cite{NRC}.  The lexicon is a list of words with their real-valued intensities for $4$ non-\textit{neutral} basic emotions. We represent the likelihood of $g$ getting mapped to each of the basic emotions, $e$, as $score_{e}($e$)$. For mapping, we first tokenize, lemmatize, and filter stop words from $g$. Then we find exact matches of $g$'s tokens to words in the lexicon (along with the match-intensities). For each match, we increase $score_{g}($e$)$ by the match-intensities. Finally, $g$ is mapped to the basic emotion with the maximum score. An emotion-phrase with no matching tokens is mapped to \textit{neutral}.

Note that we also experimented with the emotional reactions generated by $\mathbb{COMET}$ to constitute the \emph{emotion arc} without mapping them to \emph{basic emotions}. However, with more than $500$ unique emotional reactions, the space of possible arcs became too large with too few training examples for each which prevented the models from effectively learning the pattern. The smaller set of \emph{basic emotions} also made it more natural and manageable for the user to provide a desired emotion arc as input.

\subsection{Implementation Details \label{train-details}}
We follow the training and inference settings of medium-size GPT-2 as in~\citet{radford2019language} (for completion, we provide full details in Appendix~\ref{appendix:hyper-param}). Our models are implemented with the Texar toolkit~\cite{HuSTWYZHQWMLLZS19}. 
\par 

\begin{table*}[ht]
\centering
\setlength{\tabcolsep}{0.4em}
\footnotesize
\begin{tabular}{l c c c c c c c}
\toprule
 \textbf{Models} & PPL ($\downarrow$) & BLEU-1 ($\uparrow$) & BLEU-2 ($\uparrow$) & Dist-1 ($\uparrow$) & Dist-2 ($\uparrow$) & Dist-3 ($\uparrow$) & Repet-4 ($\downarrow$)\\ 
\midrule
  Fusion + Emo & 24.02 & 21.10 & 2.61 & 66.18 & 90.88 & 96.91 & 23.30\\ 
  Plan\&Write + Emo & 17.43 & 22.46 & 3.03 & 66.32 & 90.47 & 95.59 & 28.61\\ 
  PPLM-3 & -- & 20.36 & 2.37 & 71.37 & 93.90 & 98.19 & 13.36 \\
  PPLM-5 & -- & 20.61 & 2.47 & 71.47 & 93.99 & 98.21 & 14.02 \\ 

\textsc{Gpt-2} + FT* & 12.16 &  22.68 & 3.10 & 72.93 & 94.24 &  98.28 & 12.10\\ \midrule 
  EmoSup & \textbf{11.10} & 22.70 & 3.23 & 71.44 & 93.75 & 98.10 & 13.94\\
  \textsc{Rl-Em} & 11.98 & 22.52 & 3.15 & \textbf{73.32} & \textbf{94.76} & \textbf{98.56} & \textbf{10.09}\\ 
   \textsc{Rl-Clf} & 11.31 & \textbf{22.78} & \textbf{3.26} & 71.16 & 93.65 & 98.05 & 13.34\\
\bottomrule
\end{tabular}\\

\begin{tabular}{l c c c c c c}
\toprule
 \textbf{Models} & Arc-word & Seg-word & Arc-acc & Seg-acc  & \textsc{Ec-Clf} & \textsc{Ec-Em}\\ 
\midrule
  Fusion + Emo & 6.32 & 38.89 & 29.89 & 62.59 & 60.06 & 73.59\\
  Plan\&Write + Emo & 5.61 & 32.98 & 26.38 & 60.99 & 58.13 & 72.46\\ 
  PPLM-3 & 7.11 & 36.10 & 23.97 & 57.01 & 56.02 & 73.19\\
  PPLM-5 & 7.74 & 37.64 & 27.30 & 60.60 & 59.51 & 74.43\\ 

  \textsc{Gpt-2} + FT* & 4.46 & 33.28 & 17.32& 48.69 & 47.93 & 69.00 \\ \midrule
  EmoSup & 7.33 & 40.86 & 31.25 & 64.26 & 62.88 & 74.10 \\
  \textsc{Rl-Em} & 8.85 & 43.77 & 33.56 & 65.13 & 63.93 & \textbf{76.59} \\ 
  \textsc{Rl-Clf} & \textbf{10.14} & \textbf{45.42} & \textbf{37.58} & \textbf{68.90} & \textbf{67.55} & 75.87 \\
\bottomrule
\end{tabular}
\caption{\label{automatic-res}Automatic evaluation of content quality (top) and emotion faithfulness (bottom). For content quality, \textsc{Rl-Clf} and \textsc{Rl-Em} outperform all baselines for BLEU and diversity/repetition scores respectively ($p<0.05$). For emotion faithfulness, \textsc{Rl-Clf} outperforms all baselines ($p<0.05$). * indicates absence of emotion arc as input.}
\vspace{-1em}
\end{table*}

\subsection{Evaluation Measures \label{eval-met}}
\noindent\textbf{Automatic}\space\space\space We adopt several automatic measures to evaluate the generated stories both on content quality and emotion faithfulness. 

For evaluating the content quality, we use the following measures: (1) \textbf{Perplexity} as an indicator of fluency. A smaller value is better\footnote{For comparison, we compute \textit{word-level} perplexity for GPT-2 based models. That is, we normalize the total negative log probability 
by the number of \textit{word}-level tokens, not the number of BPE tokens.}.  (2) \textbf{BLEU}, which is based on $n$-gram overlaps~\cite{papineni2002bleu}. Following ~\citet{Guan2020Story}, since BLEU scores become extremely low for large $n$, we used $n{=}1,2$. (3) \textbf{Distinct-$n$} (with $n{=}1,2,3$) measure 
the percentage of unique $n$-grams~\cite{li-etal-2016-diversity}. A high ratio indicates a high level of lexical diversity. 
(4) \textbf{Repetition-4} is the percentage of generated stories that repeat at least one 4-gram~\cite{Shao2019Planning}. A high value indicates redundancy in the generated text.

For evaluating 
the emotion faithfulness of a generated story, we adapt lexical measures (1) \textbf{Seg-word} and (2)
\textbf{Arc-word}~\cite{song-etal-2019-generating}. Given a desired emotion arc for a story, Seg-word is the percentage of the story's segments that contain emotion words corresponding to desired emotion. Correspondingly, Arc-word for a story is a binary score indicating if \textit{all} of its segments contain emotion words corresponding to the desired emotions. We also define (3) \textbf{Seg-acc} and (4) \textbf{Arc-acc} for a generated story. Seg-acc for the story is the fraction of generated segments for which the emotion (as determined by the emotion classifier) exactly matches the desired emotion. Similarly, Arc-acc for a story is a binary score indicating if its emotion arc (as determined by the emotion classifier) exactly matches the desired emotion arc. We also use the reward functions, (5) \textbf{\textsc{Ec-Clf}} and (6) \textbf{\textsc{Ec-Em}}, to score a generated story. For all these measures, we report averaged scores across all stories generated by a model.

\noindent\textbf{Manual}\space\space\space 
We also conduct a manual evaluation of generated stories using Amazon Mechanical Turk. 
Following
~\citet{song-etal-2019-generating}, 
workers are asked to evaluate pair of stories on a 0-3 scale (3 being very good) from two different perspectives: (1) \textbf{emotion faithfulness} to assess whether it follows the desired emotion arc for the protagonist, and (2) \textbf{content quality} to indicate whether a story is fluent, logically coherent, and on-topic (related to the given title). Workers were also asked to indicate their \textbf{overall preference} by choosing the better story of the two while considering both aspects, or indicate that they are of equal quality.
More details about evaluation measures are provided in Appendix~\ref{appendix:eval-measures}.

\section{Results and Discussion \label{discussion}}
We first describe our experiments 
on choosing the base storytelling model (\S\ref{base-res}) followed by evaluation of the proposed models (\S\ref{main-res}).\par

\subsection{Base Storytelling Model Results\label{base-res}}
As noted before, our models build upon a base storytelling model (GPT-2). 
We compared GPT-2 with the following state-of-the-art story generation models,  given the title as input. (1) \textit{S2S}~\cite{bahdanau:14}, (2) \textit{ConvS2S} \cite{convs2s}, (3) \textit{Fusion} \cite{fan-etal-2018-hierarchical}, (4) \textit{Plan\&Write}~\cite{Yao:19}, 
(5) \textit{Decom-Event}~\cite{fan:19}.

Using various evaluation measures described earlier, our experiments showed that fine-tuned \textsc{Gpt-2} 
outperforms all baselines, indicating that it can serve as good base storytelling model. This is in-line with the observations made in \citet{Guan2020Story}. Since this is not our focus, we report 
full results and details in Appendix~\ref{appendix:base-res}. 

\subsection{Emotion-Aware Storytelling Results \label{main-res}}
\noindent\textbf{Baselines}\space\space\space We use the following baselines in our experiments: (1) \textsc{Gpt-2}+FT, our base GPT-2 model fine-tuned on the ROCStories corpus, for which emotion arcs are not provided as inputs; (2) \textit{Fusion+Emo} and (3) \textit{Plan\&Write+Emo}, which are two of the strongest storytelling baselines (we prepended emotion arcs to titles in our experiments); and (4) \textit{PPLM}~\cite{PlugPlay} which can be extended to accept emotion arcs for controlling story generation. \textit{PPLM-3} and \textit{PPLM-5} indicate $3$ and $5$ iterations respectively\footnote{We use the HuggingFace implementation: \url{https://github.com/huggingface/transformers/tree/master/examples/text-generation/pplm}. For a fair comparison, we used GPT-2 fine-tuned on stories as the underlying generation model.}.

\begin{table*}[ht]
\centering
\setlength{\tabcolsep}{0.5em}
\footnotesize
\begin{tabular}{l c c c}
\toprule
\multirow{2}[3]{*}{} & \multicolumn{2}{c}{\textbf{Specific Criteria}} & \textbf{Overall Preference} \\  \cmidrule(r){2-3} \cmidrule(l){4-4}
              &   Emotion Faithfulness & Content Quality & Better / Worse / Tie (\%)   \\
                  \midrule
\textsc{Rl-Clf} vs. \textsc{Rl-Em}  &  +0.20 \scriptsize{$(1.77\pm 0.91, 1.57\pm 0.95)$} & +0.15 \scriptsize{$(1.73\pm 0.78, 1.58\pm 0.85)$} & \textbf{52.33} / 38.00 / 9.66 \\
\textsc{Rl-Clf} vs. \textsc{Gpt-2}+FT & +0.76 \scriptsize{$(2.24\pm0.80, 1.48\pm0.98)$}  & +0.25 \scriptsize{$(2.25\pm0.82, 2.00\pm0.88)$}   & \textbf{60.00} / 22.00 / 18.00 \\ 
\textsc{Rl-Clf} vs. EmoSup & +0.28 \scriptsize{$(1.97\pm1.00, 1.69\pm1.05)$} & +0.14 \scriptsize{$(1.93\pm0.94, 1.79\pm0.97)$}  & \textbf{50.33} / 34.00 / 15.66 \\ 
\textsc{Rl-Clf} vs. PPLM-5 & +0.48 \scriptsize{$(2.10\pm 0.86, 1.62\pm 0.94)$} &  +0.34  \scriptsize{$(2.21\pm 0.90, 1.87\pm 0.96)$}  &  \textbf{61.00} / 25.66 / 13.33\\
\bottomrule
\end{tabular} %
\caption{\label{manual-res}Manual evaluation results. For each criteria, we report the average improvements as well as the absolute scores for the two models, separated by a comma. 
\textsc{Rl-Clf} is preferred over other methods ($p < 0.05$).}
\vspace{-1em}
\end{table*}

\noindent\textbf{Automatic Evaluation}\space\space\space For automatic evaluation, we used the titles and automatically extracted emotion arcs of the stories in our test set as input. 

The evaluation results on content quality are shown in the top half of Table~\ref{automatic-res}. Interestingly, even though the proposed models only aim to control emotion arc, they outperform \textsc{Gpt-2}+FT on perplexity indicating better fluency. Among the proposed models, EmoSup obtains the best perplexity score mainly because that is what its loss function optimizes 
(as opposed to the mixed loss in EmoRL models). Overall, all of our proposed models outperform all baselines. In particular, \textsc{Rl-Clf} has the highest BLEU scores, and \textsc{Rl-Em} has the highest diversity and lowest repetition scores. 
All improvements over baselines are statistically significant (approximate randomization~\cite{noreen:89}, $p<0.05$).\par

The evaluation results on emotion faithfulness are shown in the bottom half of Table~\ref{automatic-res}. We see that, as expected, all models outperform 
\textsc{Gpt-2}+FT, which is not provided the emotion arcs as inputs. Our proposed models also achieve significant improvements over all baselines (app. randomization, $p < 0.05$). In particular, \textsc{Rl-Clf} achieves the best performance on almost all measures.

\begin{figure}[t]
  \includegraphics[scale=0.47]{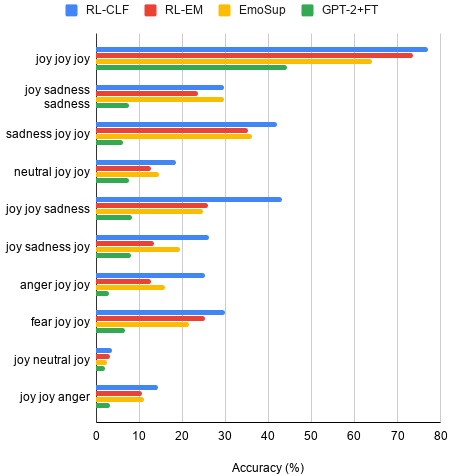} 
  \caption{Arc-acc of various models on the 10 most common arcs in our corpus. \textsc{Rl-Clf} outperforms other models for almost all arcs.}
  \label{per-arc-acc}
\end{figure}

We also compare various models on the most common emotion arcs in our corpus
. Fig.~\ref{per-arc-acc} shows the Arc-acc of various models on the $10$ most common arcs. We can see that all models perform very well on ``joy $\rightarrow$ joy $\rightarrow$ joy'' as compared to other emotion arcs. This is because this is the most common emotion arc ($34\%$ of the training data) in our corpus, which results in availability of significant number of training examples for this arc. Nevertheless, for all arcs,  \textsc{Rl-Clf} consistently outperforms all other models indicating a better control over the desired emotion arc.

These results indicate that while all proposed models can control the emotion arc of generated stories, \textsc{Rl-Clf} achieves a good balance between both content and emotion quality.

\noindent\textbf{Manual Evaluation}\space\space\space
Since concerns have been raised about automatic evaluation of language generation, we also conduct a manual evaluation on Amazon Mechanical Turk. 
For this, we randomly sampled titles and emotion arcs of $100$ instances from our test set, and generated stories using the models being evaluated. We compared five models, and so overall, there were $500$ stories. 
We conduct pairwise comparisons of generated stories, and each pair was evaluated by $3$ judges. Table~\ref{manual-res} reports the average improvements as well as absolute scores for emotion faithfulness and content quality (evaluated independently) and also the overall preference of the judges.
We first compare our two EmoRL models (Row $1$). We see that \textsc{Rl-Clf} improves over \textsc{Rl-Em} on both emotion faithfulness and content quality. Overall, it is judged to be better than \textsc{Rl-Em} $52.33\%$ of the time and worse in only $38.00\%$ cases. We then compare the better of the two, \textsc{Rl-Clf}, with the uncontrolled \textsc{Gpt-2}+FT (Row $2$). We see that on average, \textsc{Rl-Clf} model is not only better at adhering to the emotion arc by $+0.76$ points but also generates better content (improvement of $+0.25$ points) and its stories are preferred $60\%$ of the times by humans. We observe similar results for comparison with EmoSup and PPLM-5. All improvements are statistically significant (app. randomization, $p < 0.05$). 
  \par 

\begin{table}[t]
\footnotesize
\begin{tabular}{|p{1.4cm}|p{5.8cm}|}
\multicolumn{2}{l}{\textbf{Title: fire injuries}   } \\ \hline
\hlc[green]{joy} - \hlc[light-blue]{sadness} - \hlc[green]{joy}     & \hlc[green]{My friends and I went camping this summer.} We got in my van and went to the woods. We decided to light a campfire. While driving around , \hlc[light-blue]{our tire popped and the fire started.} We had to call the fire department for help and \hlc[green]{they were able to put out the fire.} \\ \hline
\hlc[light-blue]{sadness} - \hlc[light-blue]{sadness} - \hlc[green]{joy} & \hlc[light-blue]{The fire department was called to a house in the woods. The house was engulfed in flames.} There were two people inside. \hlc[light-blue]{One person was taken to the hospital by air ambulance.} \hlc[green]{Luckily, the other person was treated for non-life threatening injuries.}             \\ \hline
\multicolumn{2}{l}{\textbf{Title: dance}      } \\ \hline
\hlc[yellow]{fear} - \hlc[green]{joy} - \hlc[green]{joy}        & \hlc[yellow]{Kelly was worried about her dance recital.} \hlc[green]{She had practiced her dance for weeks.} She decided to try out for the school's dance team. Kelly was nervous but \hlc[green]{knew she could do well.} \hlc[green]{She was so excited she gave her best impression!}                                \\ \hline
\hlc[light-blue]{sadness} - \hlc[green]{joy} - \hlc[green]{joy}      & \hlc[light-blue]{I was very depressed.} \hlc[green]{I went to a dance class with a friend of mine.} We tried out some different moves. We got stuck \hlc[green]{dancing for a long time.} The next day \hlc[green]{I tried out some new moves and got a standing ovation.}                                                   \\ \hline
\end{tabular}
\caption{\label{case-study-1} For a given title, our model can generate different stories for different emotion arcs. Story segments with corresponding emotions are highlighted.}
\vspace{-1em}
\end{table}

\noindent \textbf{Case Studies}\space\space\space
Since the proposed models can generate stories conditioned on the 
protagonist's \textit{emotion arc}, they can be used to unfold a story in diverse situations for a given title. We demonstrate this capability in Table~\ref{case-study-1}. It shows two examples where for the same title, our model (\textsc{Rl-Clf} here) can generate stories that follow different emotion arcs for the protagonists.

Alternatively, given a story, the models can also be used to generate another story with a similar emotion arc (after automatically extracting the protagonist's emotion arc in the given story using the pipeline described in \S\ref{data-prep}). For example, in Table~\ref{case-study-2} we show how \textsc{Rl-Clf} can be used to generate a novel story in which the protagonist follows the same emotion arc as in the `Cinderella' story. Note that the goal here is not necessarily to generate a similar narrative but a story that follows the same emotional trajectory. \par 

We provide more qualitative examples in Appendix Figure~\ref{appendix:qualitative-ex}.

\begin{table}[t]
\footnotesize
\noindent\fbox{\begin{minipage}{\linewidth}
\textbf{Input story:} There was a girl called Cinderella who did all the work for her mean, ugly step sisters. One day, she got an invitation to go to a ball at the palace. A fairy Godmother appeared an made her a beautiful dress and a lovely carriage. After Cinderella left the ball, the prince looked everywhere for her. He eventually found her and they got married and lived happily ever after. \\
\textbf{Automatically extracted emotion arc:} \hlc[light-blue]{sadness} $\rightarrow$ \hlc[green]{joy} $\rightarrow$ \hlc[green]{joy}\\
\textbf{Input Title}: The wedding \\
\textbf{Output story}: \hlc[light-blue]{Ryan had been feeling really lonely lately.} He decided he needed a way to make a friend. \hlc[green]{He decided to go to a wedding. When he got there he met a beautiful girl.} \hlc[green]{Ryan had made a new friend that day !}
\end{minipage}}
\caption{Given a story, our model can generate another story with similar emotion arc.}
\label{case-study-2} 
\vspace{-1em}
\end{table}

\section{Conclusion}
In this paper, we proposed the emotion-aware storytelling task for modeling the emotion arc of the protagonist. To this goal, we designed two emotion-consistency rewards using a commonsense transformer and an emotion classifier. 
Experiments demonstrated that our approach improved both content quality and emotion faithfulness of the generated stories. 
We also presented two case studies, which show interesting use cases of our model. In general, such models can have educational applications by enabling children to explore creative writing at an early age and addressing the literary learning needs of learners with disabilities.

This paper is a step towards future research directions on planning emotional trajectory while generating stories. Using commonsense inferences about the effect of the 
events on emotional states of various characters of the story has the potential of generating more coherent, realistic, and engaging stories.  
In this work, we focused only on the protagonist, but future works can explore modeling motivations, goals, achievements, and emotional trajectory of all characters. Our approach is general and provides a blueprint for similar works going forward and can be used outside emotion-aware storytelling, e.g., for generating other emotional content or text with other attributes or properties.

The various assumptions and choices made in this paper and  the specific characteristics of the dataset we chose can introduce biases and errors. For example, $\mathbb{COMET}$ is a discourse-agnostic model, and separately extracting emotional reactions for each sentence may fail to maintain emotional consistency with the rest of the narrative. Such sources of errors and biases need further systematic investigation.

\bibliography{emnlp2020}
\bibliographystyle{acl_natbib}

\newpage

\appendix
\input{appendix_camera}
\label{sec:appendix}

\end{document}

%% file: figures/example.tex
\begin{figure}[t]
 \centering
 \footnotesize
\noindent\fbox{\begin{minipage}{\linewidth}
\textbf{Title (input):} Raw burger \\
\textbf{Emotion arc (input):} \hlc[green]{joy} $\rightarrow$ \hlc[pink]{anger} $\rightarrow$ \hlc[light-blue]{sadness}\\
\textbf{Story (output)}: \hlc[green]{Tom went to a burger place with his friends.} He ordered a burger. When he got it , he noticed that it was raw. \hlc[pink]{Tom yelled at the waiter} for it being raw. \hlc[light-blue]{He was really disappointed.}
\end{minipage}}

\caption{An example story generated by our model for a given title and emotion arc of the protagonist. Story segments  
are highlighted with the emotions the protagonist (Tom) experiences.}

\label{example}
\end{figure}

%% file: appendix_camera.tex
\section{Training and Experiment Setup}


\subsection{Obtaining Emotion Reactions During \textsc{Rl-Em} Training}
\label{appendix:label-training}
During self-critical training, for each training instance, two stories are generated: $\bm{y}^s$ which is sampled from the model's probability distribution, and $\hat{\bm{y}}$ which is greedily decoded. Then for each sampled or greedy story, the value of the reward is computed. For computing the \textsc{Ec-Em} reward, we need to identify and track the protagonist and obtain his/her emotional reactions for each sentence of the sampled or greedy story. This requires identifying the protagonist and determining his/her role, \textit{Agent} or \textit{Other}, in every sentence so that the appropriate argument for $\mathbb{COMET}$ (\texttt{xReact} or \texttt{oReact}) can be chosen. In principle, this can be done using the annotation pipeline described in \S4.1 of the main paper. However, doing this is computationally prohibitive during training as the pipeline requires running dependency parsing for each sentence and coreference resolution for every sampled or greedy story. 
To this end, upon analyzing our corpus and some generated stories, we devised several heuristics that approximate the tasks (i.e. identifying protagonists and their roles) with high accuracy. We describe these heuristics below. 

For identifying the protagonist, we use the following heuristics. The first heuristic is based on the observation that if the narrator features in a story, the story primarily focuses on the narrator and  his/her experiences, thus making him/her the protagonist. So our first heuristic is that if first person pronouns (I, We) appear in the story, they are considered the protagonist.  Our second heuristic is based on the observation that the protagonist is usually introduced fairly early in the story. Especially, in our case, where the stories are 5-sentence long, the protagonist appears mostly in the first couple of sentences of the story. With this in mind, we define the first noun that appears in a lexicon of common protagonists as the protagonist of the story. This lexicon consists of terms for \texttt{Male\_Char}, \texttt{Female\_Char}, \texttt{Social\_Group}, \texttt{Generic\_People}, and the NLTK name corpus. Example of these terms are shown in Table~\ref{names}. This also let's us identify the gender of the protagonist (using the lexicon category that the protagonist belongs to) and hence the pronouns that will be used in the following sentences to refer to the protagonist. This combination of protagonist's mentions and pronouns lets us track him/her throughout the story.  

For identifying the role of the protagonist, if the first noun that occurs in a sentence matches the protagonist or his/her corresponding pronoun, we assume that the protagonist's role is \textit{Agent}, otherwise the role is \textit{Other}. Depending on this role, we use \texttt{xReact} or \texttt{oReact} when obtaining protagonist's emotional reaction in that sentence using $\mathbb{COMET}$.

\begin{table}[!ht]
\scriptsize
\def\arraystretch{1.5}
	\begin{center}
	    \begin{tabularx}{\linewidth}{@{\hspace{0pt}}p{1.8cm}X@{\hspace{0pt}}}
	\hline \toprule[0.65pt]
	
	\textbf{\texttt{Male\_Char}} & husband, father, dad, dady, brother, grandpa, granddad, son, nephew, man, boy, boyfriend\\\hline
	\textbf{\texttt{Female\_Char}} &  wife, mother, mom, momy, sister, grandma, grandmom, niece, daughter, nana, woman, girl, girlfriend \\ \hline 
	\textbf{\texttt{Social\_Group}} & family, parents, grandparents, children, kids, couple, friends, boys, girls, band \\ \hline
	\textbf{\texttt{Generic\_People}} &  cousin, friend, fiance, boss, manager, assistant, doctor, nurse\\\hline 
	
\end{tabularx}
	\end{center}
	\caption{\label{names} Predefined terms used for tracking the protagonist.}
	\vspace{-1em}
\end{table}

\subsection{Training Hyper-parameters}
\label{appendix:hyper-param}
Our proposed models follow the setting of medium-sized GPT-2~\cite{radford2019language} ($345$ million parameters) that used a $24$-layer decoder-only transformer, $1024$-dimensional hidden states, and $16$ attention heads. The stories are encoded using BPE with vocabulary size of $50,257$. We set the maximum sequence length to $128$ tokens, as it is large enough to contain complete stories and additional inputs. We use Adam optimization~\cite{adam:14} with an initial learning rate of $10^{-5}$ and minibatch of size $4$.  For stability, we first pre-train the models with teacher forcing until convergence, then fine-tune them with the mixed loss. 
Hyper-parameter $\gamma=0.97$ is tuned manually on the validation set. All models were trained until there was no improvement on the validation set performance. We use a NVIDIA GTX 1080 Ti GPU machine 
to train our models. At inference time, we generate stories using top-$k$ sampling scheme~\cite{fan-etal-2018-hierarchical} with $k{=}40$ and a softmax temperature of $0.7$. It took about $3$ hours to generate stories for our test set of size $9,796$.\par 
For generating commonsense inferences about the protagonist's emotions using $\mathbb{COMET}$, we use greedy decoding algorithm since it has been shown to have superior performance as evaluated by humans~\cite{Bosselut2019COMETCT}.

\begin{table}[t]
\centering
\scriptsize
\setlength{\tabcolsep}{0.25em}
\small
\begin{tabular}{l c c c c}
\toprule
\vspace{-0.25em}
\textbf{Models} & Domain & \thead{Accuracy \\(Jaccard)} & Mico F1  & Macro F1\\ 
\midrule
  \citet{meisheri-dey-2018-tcs} & tweets & 0.582 & 0.694 & 0.534\\
  \citet{Baziotis} & tweets & 0.595 & 0.709 & 0.542\\
 \citet{kant2018practical} & tweets & 0.577& 0.690 & 0.561\\
 BERT$_{\textit{large}}$ (ours) & tweets & 0.595 & 0.708 & 0.522\\ \hline
 BERT$_{\textit{large}}$ (ours) & stories & 0.617 & 0.650 & 0.557 \\
\bottomrule
\end{tabular}
\caption{\label{clf-result} Emotion classification results on the tweets dataset (upper block), and the automatically annotated story corpus (lower block). }
\vspace{-1em}
\end{table}

\subsection{Evaluation Measures}
\label{appendix:eval-measures}
In this section we provide details about the automatic and manual measures used to evaluate our models. 

\textbf{Automatic}\space\space\space To compute Arc-word and Seg-word measures, we use NRC Affect Intensity Lexicon~\cite{NRC}. This lexicon contains words with corresponding emotion-intensities for different \textit{basic emotions}. To find emotionally expressive words in a given piece of text (e.g., a story segment), we create a dictionary of words with emotion intensity higher than $0.5$ for each of our \textit{basic emotions}. \par

\begin{figure*}
\centering
  \includegraphics[scale=0.45]{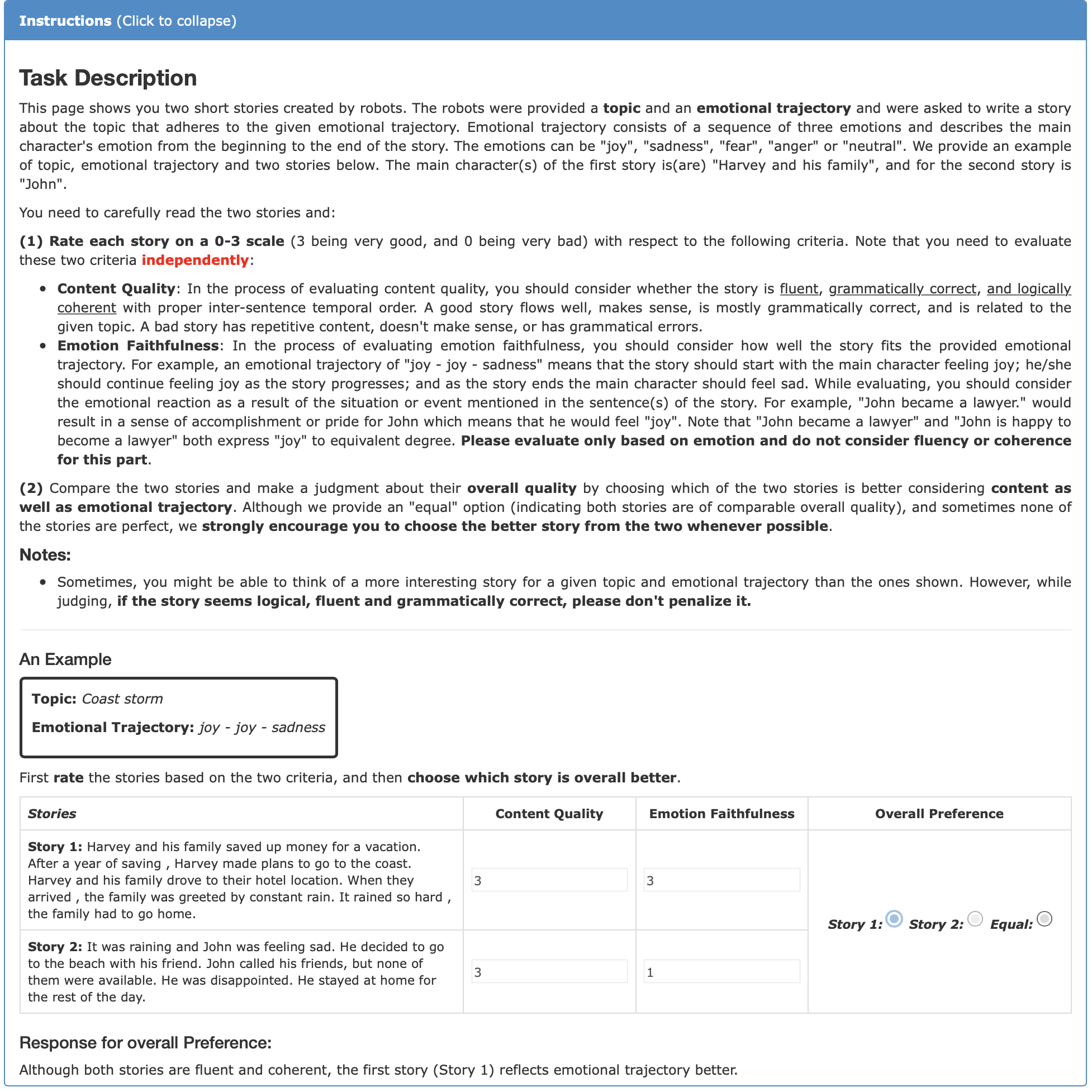} 
  \caption{An screenshot of manual evaluation on AMT.}
  \label{AMT-shot}
\end{figure*}

\noindent\textbf{Manual}\space\space\space During the manual evaluation, we conducted pairwise comparison of the models on Amazon Mechanical Turk (AMT). To ensure high quality of evaluation, we selected turkers that  had an approval rate greater than $97\%$, had at least $1,000$ approved HITS, and were located in the U.S. For each pairwise annotation, we showed the inputs (title and emotion arc) and two stories generated using the two models being compared. In order to avoid biases, we randomly shuffled the order in which the stories from the two models were shown to the turkers. We provided instructions to the turkers explaining the annotations and also provided examples. Following this process, each pair of stories was annotated by three turkers. Fig.~\ref{AMT-shot} shows a screenshot of our  setup on AMT.

\begin{figure}
\centering
  \includegraphics[scale=0.45]{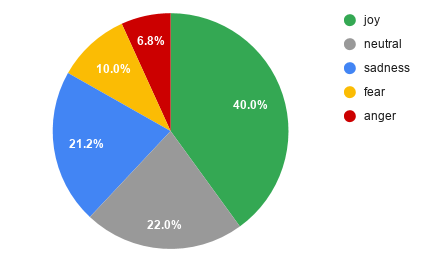} 
  \caption{Label distribution of human annotated set.}
  \label{dist-chart}
\end{figure}

\section{Supplementary Results}

\subsection{Emotion Classification \label{appendix:clf-res}}

Our \textsc{Ec-Clf} reward captures the protagonist's emotions using our emotion classifier. In this section we provide details about its evaluation. 

We first evaluate the classifier on the tweets corpus~\cite{mohammad-etal-2018-semeval}\footnote{\url{https://competitions.codalab.org/competitions/17751}} by comparing it with several strong baselines~\cite{kant2018practical}. For this comparison, we trained all models on the training set of the corpora and tested them on a held-out test set. The models were evaluated using Jaccard Index based accuracy, and Micro and Macro F1 scores. This evaluation set-up (train-validation-test splits and choice of evaluation metrics) is as suggested in the challenge that provided the corpus (SemEval Task1:E-c challenge). The results of this comparison is shown in the top half of Table~\ref{clf-result}. We can see that our emotion classifier, BERT$_{\textit{large}}$, is superior or competitive with other models.

The results reported above show that the model performs well for emotion classification in tweets. However, our goal is to design a model that can be used to track protagonist's emotions in stories. As described in the main paper, we further fine-tuned this classifier on our automatically annotated story corpora (described in the paper in \S\ref{data-prep}). We also evaluated the classifier on a held-out portion of this story corpora consisting of about $1,201$ stories ($6,005$ sentences in total).
The results are reported in the last row of Table~\ref{clf-result}. The classifier achieves a (Jaccard Index) accuracy of $61.75\%$ and micro and macro F1 scores of $0.650$ and $0.557$ respectively. Note that this is different from the evaluation reported in the paper, which was conducted on a subset of stories annotated by humans.

\subsection{Manual Annotation for Protagonist's Emotions}
\label{appendix:manual-annot}
\begin{figure*}
\centering
  \includegraphics[scale=0.45]{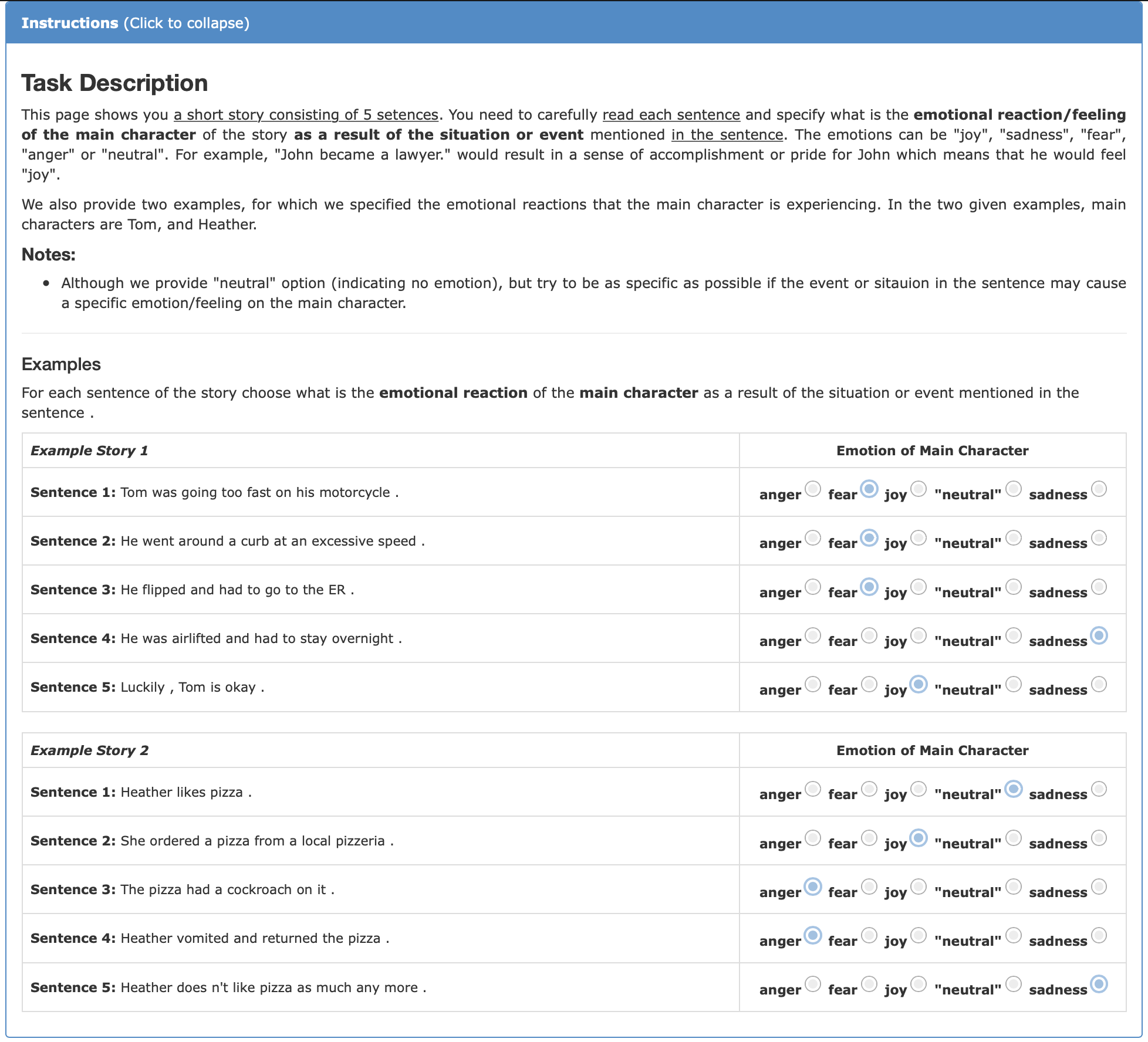} 
  \caption{A screenshot of human annotation for protagonist's emotions.}
  \label{AMT-CLF}
\end{figure*}

As described in the paper (\S\ref{comet}), the emotion classifier was also evaluated on a subset of $50$ randomly selected stories ($250$ sentences) manually annotated for the emotions experienced by their protagonists. This annotation was done on Amazon Mechanical Turk. To ensure good quality of the annotations, we selected turkers who had an approval rate greater than $97\%$,  had at least $1,000$ approved HITS, and were located in the U.S. Fig.~\ref{AMT-CLF} shows a screenshot of our setup. The Fleiss kappa (inter-annotator agreement) was also moderate ($\kappa=0.55$). We also analyzed the annotations to identify major sources of disagreements between the turkers. We found that most disagreements occurred between \textit{neutral} and \textit{joy}; and also between \textit{sadness} and \textit{anger}. The overall label distribution for this human annotated set is shown in Fig~\ref{dist-chart}.

\begin{table*}[ht]
\centering
\setlength{\tabcolsep}{0.4em}
\footnotesize
\begin{tabular}{l|c c c c c c c}
\toprule
 \textbf{Models} & PPL ($\downarrow$) & BLEU-1 ($\uparrow$) & BLEU-2 ($\uparrow$) & Dist-1 ($\uparrow$) & Dist-2 ($\uparrow$) & Dist-3 ($\uparrow$) & Repet-4 ($\downarrow$)\\
\midrule
  S2S & 22.60 & 21.98 & 2.82 & 64.00 & 87.07 & 93.90 & 36.57\\
  ConvS2S & 23.90 & 20.37 & 2.63 & 66.98 & 91.22 & 95.68 & 19.29\\ 
  Fusion & 22.34 & 20.44 & 2.66 & 67.05 & 91.26 & 95.71 & 19.26\\ 
  Plan\&Write & 21.11$\ssymbol{2}$ & 22.27 & 3.03 & 65.18 & 88.45 & 95.11 & 30.43\\ 
  Decom-Event & 18.16$\ssymbol{2}$ & 18.14 & 1.69 & 68.82 & 92.39 & 97.40 & 20.52\\ \midrule
 \textsc{Gpt-2} + FT & \textbf{12.16} &  \textbf{22.68} & \textbf{3.10} & \textbf{72.93} & \textbf{94.24} &  \textbf{98.28} & \textbf{12.10}\\ 
\bottomrule
\end{tabular}
\caption{\label{base-model-comp} Base Storytelling model: Automatic evaluation. The scores marked with $\ssymbol{2}$ indicate models that have access to extra ground-truth information besides title (keywords and event tuples).}
\vspace{-1em}
\end{table*}

\subsection{Base Storytelling Model}
\label{appendix:base-res}
\textbf{Baselines}\space\space We compare our base storytelling model, GPT-2, with following state-of-the-art models:
\begin{enumerate} 
    \item \textit{S2S}, an LSTM-based seq2seq model with attention~\cite{bahdanau:14}
    \item \textit{ConvS2S}, a convolutional seq2seq model~\cite{convs2s} with decoder self-attention.
    \item \textit{Fusion}, a storytelling model that first pretrains a convolutional seq2seq model, then fixes the trained model and passes it to the second clone model with fusion mechanism~\cite{fan-etal-2018-hierarchical}.
    \item \textit{Plan\&Write}, another storytelling model that, given a title,  first generates a plot as a sequence of keywords, and then conditioned on the plot it generates the text of the story~\cite{Yao:19}.
    \item \textit{Decom-Event}, \citet{fan:19} proposes to decompose the generation to two steps: generating successive events as the story plots, and generating story by surface realization of the plots. Events are represented by 4-tuples ${<}s, v, o, m{>}$, where $s$ and $o$ are the subject and object of verb $v$, and $m$ is the modifier.
\end{enumerate} 
All of these models are trained, validated and tested on the same data splits described in \S\ref{data-prep}.

\textbf{Results}\space\space The models are evaluated on content quality using the automatic measures described in the paper in \S\ref{eval-met}. For comparison with other baselines, we compute \textit{word-level} perplexity (PPL) for \textsc{Gpt-2}+FT. That is, we normalize the total negative log probability of the target text by the number of \textit{word}-level tokens (similar to the baselines), not the number of BPE tokens. \par 

Table~\ref{base-model-comp} describes the results of this evaluation. We can see that fine-tuned \textsc{Gpt-2} performs better than all baselines for all measures. 
This demonstrates that it can be used as a good base storytelling model upon which our models are built.  

\subsection{Emotional-Aware Storytelling}
\label{appendix:emo-aware-more-res}

\noindent \scalebox{.95}[1.0]{\textbf{Supplementary qualitative examples}}\space\space\space We provide more qualitative examples in Fig~\ref{appendix:qualitative-ex}. In the figure we show stories generated by our model for a given title and emotion arc for the protagonist.


\begin{figure}[ht]
 \centering
 \footnotesize

\noindent\fbox{\begin{minipage}{\linewidth}
\textbf{Title:} A new outfit \\
\textbf{Emotion arc:} \hlc[green]{joy} $\rightarrow$ \hlc[green]{joy} $\rightarrow$ \hlc[green]{joy}\\
\textbf{Story}: \hlc[green]{Sara went to the mall.} She wanted to find something cute to wear. \hlc[green]{She noticed a beautiful red dress she liked. Sara bought the dress.} \hlc[green]{She was happy that she found something nice.}
\end{minipage}}

\noindent\fbox{\begin{minipage}{\linewidth}
\textbf{Title:} Wedding \\
\textbf{Emotion arc:} \hlc[green]{joy} $\rightarrow$ \hlc[pink]{anger} $\rightarrow$ \hlc[green]{joy}\\
\textbf{Story}: \hlc[green]{Ben is getting married. He is having fun.} \hlc[pink]{Suddenly , Ben gets into a huge fight with a guy.} Ben gets out of the wedding early. \hlc[green]{Ben is so happy to have his wedding over.}
\end{minipage}}

\noindent\fbox{\begin{minipage}{\linewidth}
\textbf{Title:} Study\\
\textbf{Emotion arc:} \hlc[yellow]{fear} $\rightarrow$ \hlc[yellow]{fear} $\rightarrow$ \hlc[light-blue]{sadness}\\
\textbf{Story}: \hlc[yellow]{Tina was so nervous for her big exam.} She did not study at all. \hlc[yellow]{She was so afraid she would fail.} She ended up failing. \hlc[light-blue]{Tina felt very sad.}
\end{minipage}}

\noindent\fbox{\begin{minipage}{\linewidth}
\textbf{Title:} The big storm \\
\textbf{Emotion arc:} \hlc[green]{joy} $\rightarrow$ \hlc[green]{joy} $\rightarrow$ \hlc[light-blue]{sadness}\\
\textbf{Story}: \hlc[green]{It was a beautiful day in the city.} \hlc[green]{The sky was clear and the weather was mild. The school was hosting an evening celebration.} A big storm rolled through and the students were worried. \hlc[light-blue]{The storm destroyed all the decorations and ruined the party.}
\end{minipage}}

\noindent\fbox{\begin{minipage}{\linewidth}
\textbf{Title:} Delayed flight \\
\textbf{Emotion arc:} \hlc[green]{joy} $\rightarrow$ \hlc[pink]{anger} $\rightarrow$ \hlc[light-blue]{sadness}\\
\textbf{Story}: \hlc[green]{Tom was going on a vacation.} He decided to go to Florida. Unfortunately his flight was delayed. \hlc[pink]{He was so frustrated he called his airline.} \hlc[light-blue]{His airline cancelled his flight.}
\end{minipage}}

\noindent\fbox{\begin{minipage}{\linewidth}
\textbf{Title:} The new pet \\
\textbf{Emotion arc:} \hlc[gray]{neutral} $\rightarrow$ \hlc[green]{joy} $\rightarrow$ \hlc[green]{joy}\\
\textbf{Story}: \hlc[gray]{Sam was walking around the neighborhood.} \hlc[green]{She saw a cute little dog.} She decided to take him home. \hlc[green]{He got along well with everyone.} \hlc[green]{Sam was glad to have a companion.}
\end{minipage}}

\noindent\fbox{\begin{minipage}{\linewidth}
\textbf{Title:} Larry practice yoga \\
\textbf{Emotion arc:} \hlc[yellow]{fear} $\rightarrow$ \hlc[green]{joy} $\rightarrow$ \hlc[green]{joy}\\
\textbf{Story}: \hlc[yellow]{Larry has always felt nervous about yoga.} He has tried many times to practice but has never gotten the hang of it. He decides to take a yoga class at his local yoga studio. \hlc[green]{He is amazed by the benefits and feels confident about his yoga practice.} \hlc[green]{Larry is happy he learned to enjoy yoga.}
\end{minipage}}

\caption{Qualitative examples of generated stories given a title and an emotion arc. }
\label{appendix:qualitative-ex}
\end{figure}